\documentclass[journal, onecolumn]{IEEEtran}
\ifCLASSINFOpdf
   \usepackage[pdftex]{graphicx}
   \usepackage[skip=2pt, font=footnotesize, labelsep=period]{caption}
   \usepackage{multirow}
   \usepackage{tabularew}
   \usepackage[flushleft]{threeparttable}
   \usepackage{subfig}
\else
\fi
\usepackage[cmex10]{amsmath}
\IEEEoverridecommandlockouts

\makeatletter
\renewcommand{\p@subfigure}{\thefigure(}
\makeatletter
\renewcommand{\figurename}{Figure}
\renewcommand\fnum@figure{\textbf{\figurename\nobreakspace\thefigure}}

\usepackage{etoolbox}
\patchcmd{\section}{\centering}{}{}{}

\renewcommand\thetable{\arabic{table}}
\renewcommand\tablename{Table}
\renewcommand\fnum@table{\textbf{\tablename\nobreakspace\thetable}}

\begin{document}
\title{Proposal for a Leaky-Integrate-Fire Spiking Neuron based on Magneto-Electric Switching of Ferro-magnets}

\author{\IEEEauthorblockN{Akhilesh Jaiswal, Sourjya Roy, Gopalakrishnan Srinivasan and Kaushik Roy\\}
\IEEEauthorblockA{School of Electrical and Computer Engineering,
Purdue University, West Lafayette, US\\
Email: \{jaiswal, roy48, srinivg, kaushik\}@purdue.edu}

%\\[-3.0ex]
\vspace{-2.0ex}
}


%Telephone: (800) 555--1212, Fax: (888) 555--1212}


\maketitle
\thispagestyle{plain}
\pagestyle{plain}

\begin{abstract}
%\boldmath
\normalsize
The efficiency of the human brain in performing classification tasks has attracted considerable research interest in brain-inspired neuromorphic computing. Hardware implementations of a  neuromorphic system aims to mimic the computations in the brain through interconnection of \textit{neurons} and \textit{synaptic} weights. A leaky-integrate-fire (LIF) spiking model is widely used to emulate the dynamics of neuronal action potentials. In this work, we propose a spin based LIF spiking neuron using the magneto-electric (ME) switching of ferro-magnets. The voltage across the ME oxide exhibits a typical leaky-integrate behavior, which in turn switches an underlying ferro-magnet. Due to the effect of thermal noise, the ferro-magnet exhibits probabilistic switching dynamics, which is reminiscent of the stochasticity exhibited by biological neurons. The energy-efficiency of the ME switching mechanism coupled with the intrinsic non-volatility of ferro-magnets result in lower energy consumption, when compared to a CMOS LIF neuron. A device to system-level simulation framework has been developed to investigate the feasibility of the proposed LIF neuron for a hand-written digit recognition problem.
\\
\end{abstract}
\setlength{\textfloatsep}{5pt}


% \fi
\IEEEpeerreviewmaketitle

%%\begin{IEEEkeywords}
%%Spiking Neural Networks, Binary Synapse, Long-Term Short-Term Memory, Stochastic STDP, Magnetic Tunnel Junction, spin-Hall Effect.
%%\end{IEEEkeywords}

\vspace{-3.5ex}

\section*{\large\bf{Introduction}}
%\IEEEPARstart{T}{he}

\begin{figure}[h]
\includegraphics[width=7in]{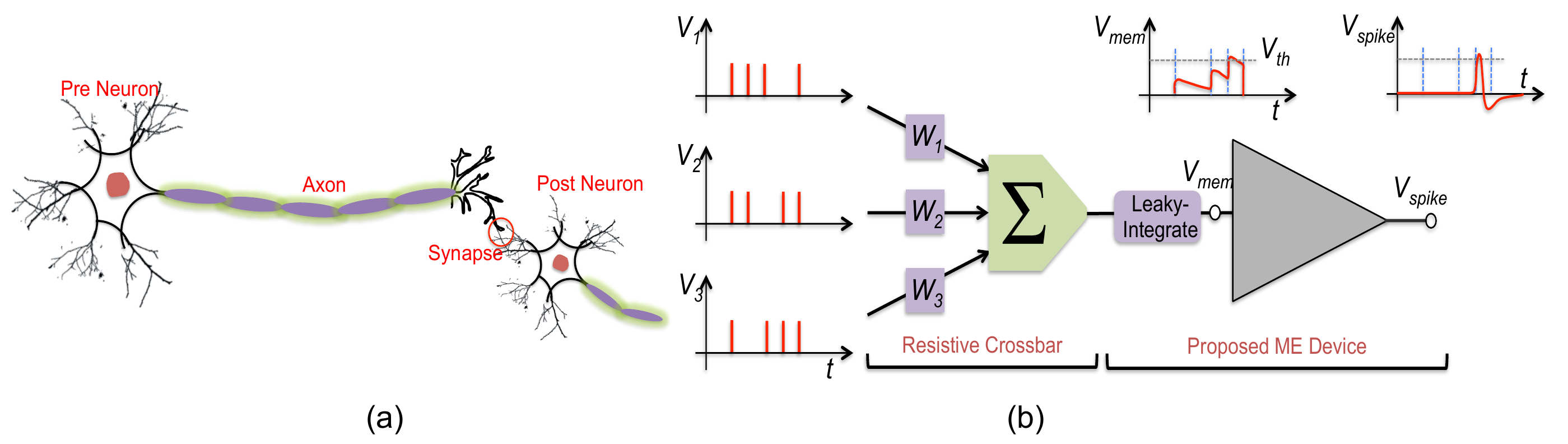}% Here is how to import EPS art
\caption{\label{fig:epsart} (a) A biological neuron. Upon sufficient electrical excitation, the input neuron called the pre-neuron communicates with the output neuron called the post-neuron by sending electrical spikes along its axon. The connection between the pre-neuron and the post-neuron is called a synapse. The spike generated by the input neuron is modulated by the strength of the synaptic connection. Further, learning occurs by altering the strength (weight) of the synapse based on the timing of the spikes generated by the pre-neuron and post-neuron.  (b) A representative model for a biological neural network. $V_i$s are the input spikes generated by pre-neurons. These input spikes are modulated by the weights $W_i$s and summed up together. The summation output alters the membrane potential of a leaky-integrate-fire neuron. The neuron emits a spike, if the membrane potential ($V_{mem}$) crosses a certain threshold ($V_{th}$). For hardware implementation, the weighted summation is usually carried out by a resistive crossbar array. Our proposed ME device aims to emulate the LIF and thresholding behavior of a biological neuron.}
\end{figure}

The brain-inspired spiking neural network (SNN) is composed of a set of \textit{pre} (input) and \textit{post} (output) neurons connected through synapses, as illustrated in Fig. 1(a). Upon sufficient excitation, the neurons generate \textit{action potentials} called \textit{spikes}, and encode information in the timing or frequency of the spikes. The interconnecting synaptic weights are updated in correlation to the timing of the spikes generated from the pre- and post-neuron. Fig. 1(b), shows a widely accepted simplified model for biological neural networks. The input spike trains are generated by the pre-neurons. The spikes from all the pre-neurons are altered as per the associated weights $W_i$, and summed up as shown in the figure. The output after summation, alters the \textit{membrane potential ($V_{mem}$)} of the post-neuron in a typical leaky-integrate fashion \cite{LIF_neuron}, as shown in inset in Fig. 1(b). If the \textit{membrane potential} crosses a certain threshold ($V_{th}$), the post-neuron emits a spike.

On-chip SNNs are being extensively explored in order to emulate the energy-efficiency of the human brain for classification applications. As opposed to a CMOS implementation, non-volatile devices mimicking neuronal characteristics are well suited for such a sparse system like SNN due to low leakage power consumption. Typically, non-volatile resistive crossbar networks are used to store the weights and carry out the summation operation. However, the analog LIF dynamics of a biological neuron are generally implemented in an area in-efficient and power-expensive CMOS technology. In this paper, we propose a novel non-volatile spin based LIF neuron using the ME switching of ferro-magnets. The key contributions of the present work are as follows:

1) We explore a spin based LIF spiking neuron based on an ME oxide in contact with an underlying ferro-magnet. The dynamics of the accumulated volatge across the ME oxide constitutes the required leaky-integrate behavior. When a sufficient voltage develops accross the ME oxide the underlying ferro-magnet switches to generate a spike.   

2) A coupled numerical simulation framework, including magnetization dynamics governed by stochastic Landau-Lifshitz-Gilbert equation and non-equilibrium Green's function (NEGF) based transport model, has been developed for analyzing the proposed ME neuron. 

3) Further, using a device to system-level simulation methodology, we have trained an SNN to recognize handwritten digits from a standard dataset. Owing to the LIF characteristic of the proposed neuron, a classification accuracy close to 70\% is achieved for 100 excitatory post-neurons.

\section*{\large\bf{Proposed Magneto-Electric Neuron}}

\begin{figure}[h]
\centering
\includegraphics[width=4in]{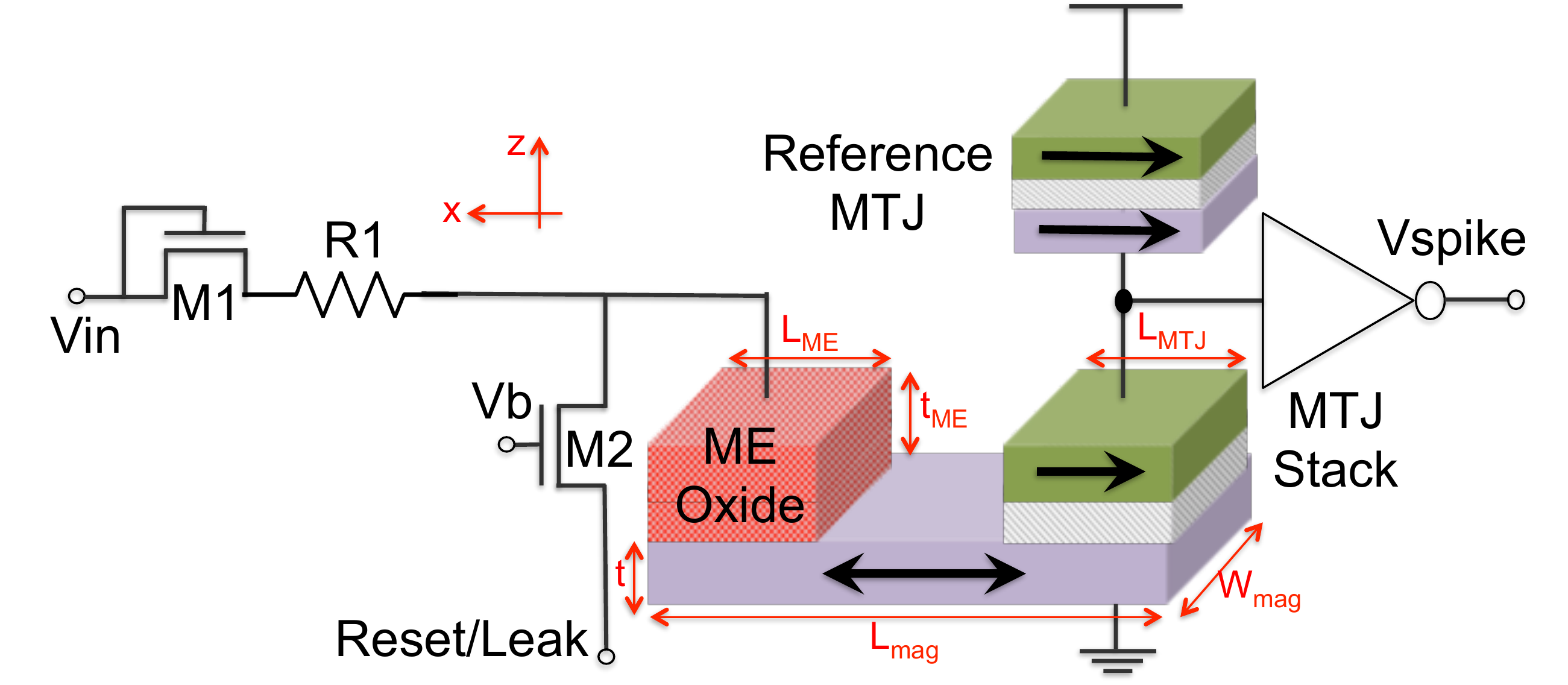}% Here is how to import EPS art
\caption{\label{fig:epsart} Schematic of the proposed LIF ME neuron. Thick ME oxide (5nm) sandwiched between the metal contact and the ferro-magnet, acts as a capacitor. Diode connected transistor M1 prevents back flow of charges stored on the ME capacitor, while resistor R1 determines the rising time constant for the capacitor. M2 constitutes the leak path, when the voltage on the Leak/Reset terminal is zero. Whereas a negative voltage at the Leak/Reset terminal, resets the ferro-magnet to its initial state (+x direction). At the output, a reference MTJ and an inverter is used to generate a voltage spike based on the state of the ferro-magnet beneath the ME oxide. }
\end{figure}

Magneto-electric effect is the physics of induction of magnetization in response to an electric field. ME effects have been experimentally demonstrated in intrinsic as well as composite multi-ferroics \cite{ME_renais}. The microscopic origin of the ME effect can be either due to exchange coupling, strain coupling or anisotropy change. To the first order, irrespective of the origin, the ME effect is linear and can be abstracted in a parameter called $\alpha_{ME}$ \cite{ME_ramesh}. $\alpha_{ME}$ is the ratio of magnetic field generated per unit electric field. Experimentally, values of $\alpha_{ME}$ have been demonstrated upto 1x10$^{-7}sm^{-1}$ \cite{ME_ramesh}. If the magnetic field, thus generated by application of an electric voltage, is strong enough, it can switch an underlying ferro-magnet.

Based on the aforementioned ME effect, the proposed neuronal device is shown in Fig. 2. It consists of a ferro-magnet under a thick ME oxide like BiFeO$_3$. The metal contact to the ME oxide and the underlying ferro-magnet form two plates of the ME capacitor. In this work, we assume a positive voltage on ME capacitor switches the ferro-magnet in -x direction and a negative voltage switches it to +x direction. 

The ferro-magnet is extended to form the free layer for an MTJ (magnetic tunnel junction), consisting of the usual \textit{free layer} - MgO (Tunnel oxide) - \textit{pinned layer} stack. The reference MTJ and the bottom MTJ form a voltage divider. The ferro-magnet under the ME oxide is initially reset to +x direction by applying a negative pulse on the \textit{Leak/Reset} terminal shown in Fig. 2. 

After the reset phase, the \textit{Leak/Reset} terminal is set to zero volts. Therefore transistor M2 acts as a leak path for the ME capacitor. On the other hand, diode connected M1 and R1 constitute the charging path. Thus, the voltage on the ME capacitor follows the leak and integrate dynamics of a biological neuron due to co-existence of a charging (M1-R1) and discharging (M2) path. If the ME capacitor is sufficiently charged, such that the generated magnetic field is greater than the anisotropy field of the ferro-magnet, the magnet switches from its initial reset position (+x direction) to -x direction, thus mimicking the thresholding behavior of a biological neuron. As the ferro-magnet switches to -x direction, the MTJ stack that has its pinned layer always pointing in -x direction, transitions from high resistance (anti-parallel) to low resistance (parallel) state. Thereby, due to the voltage divider effect, the output of the inverter goes from low to high, thereby generating an output spike.

\section*{\large\bf{Device Modeling and Simulation}}

\begin{figure}[h]
\centering
\includegraphics[width=6in]{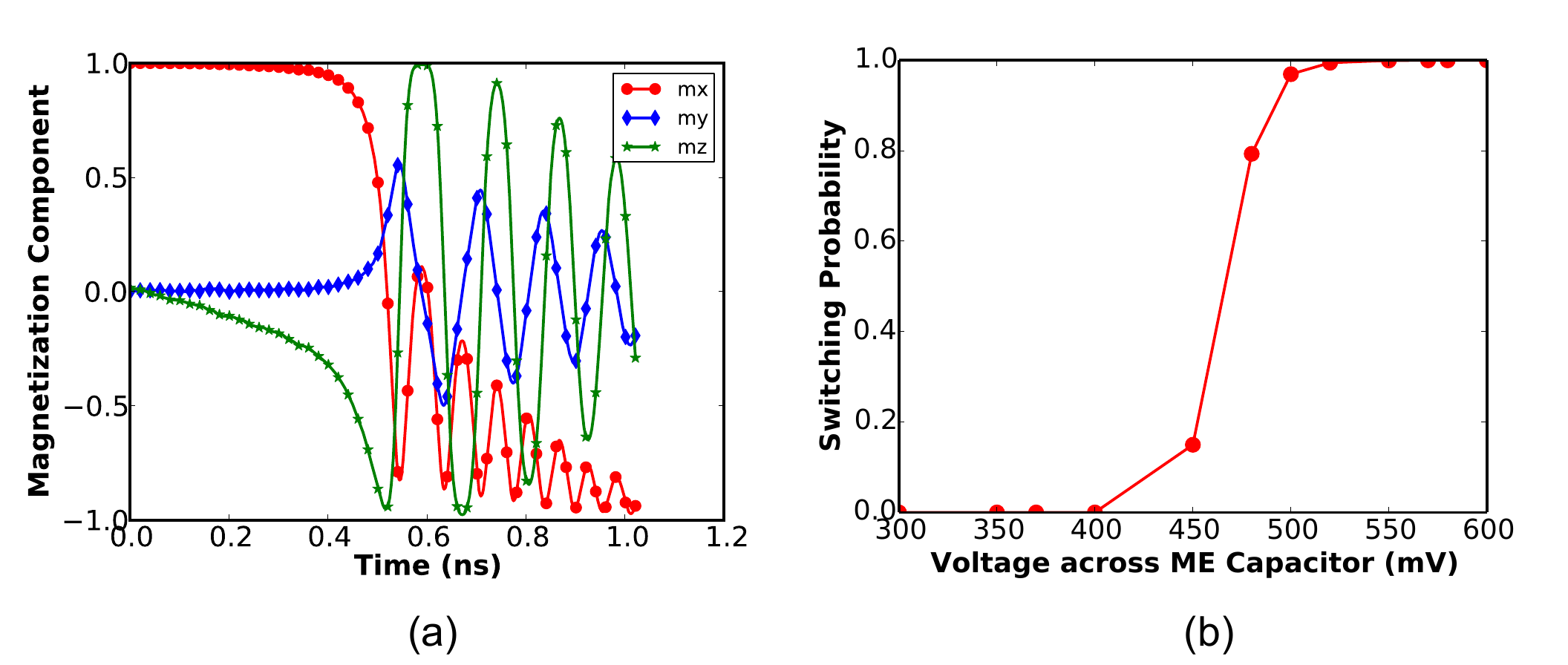}% Here is how to import EPS art
\caption{\label{fig:epsart} (a) Typical evolution of magnetization components, under influence of an electric field, when the magnet switches from +x to -x direction. (b) The stochastic switching behavior of the proposed ME neuron as a function of the voltage across ME capacitor. The switching probability was obtained for 10,000 runs using our magnetization dynamics model with thermal noise and pulse duration of 1ns. The stochastic switching dynamics of the proposed ME device is desirable, since biological neurons are known to be stochastic in their behavior.}
\end{figure}

\begin{figure}[h]
\centering
\includegraphics[width=7in]{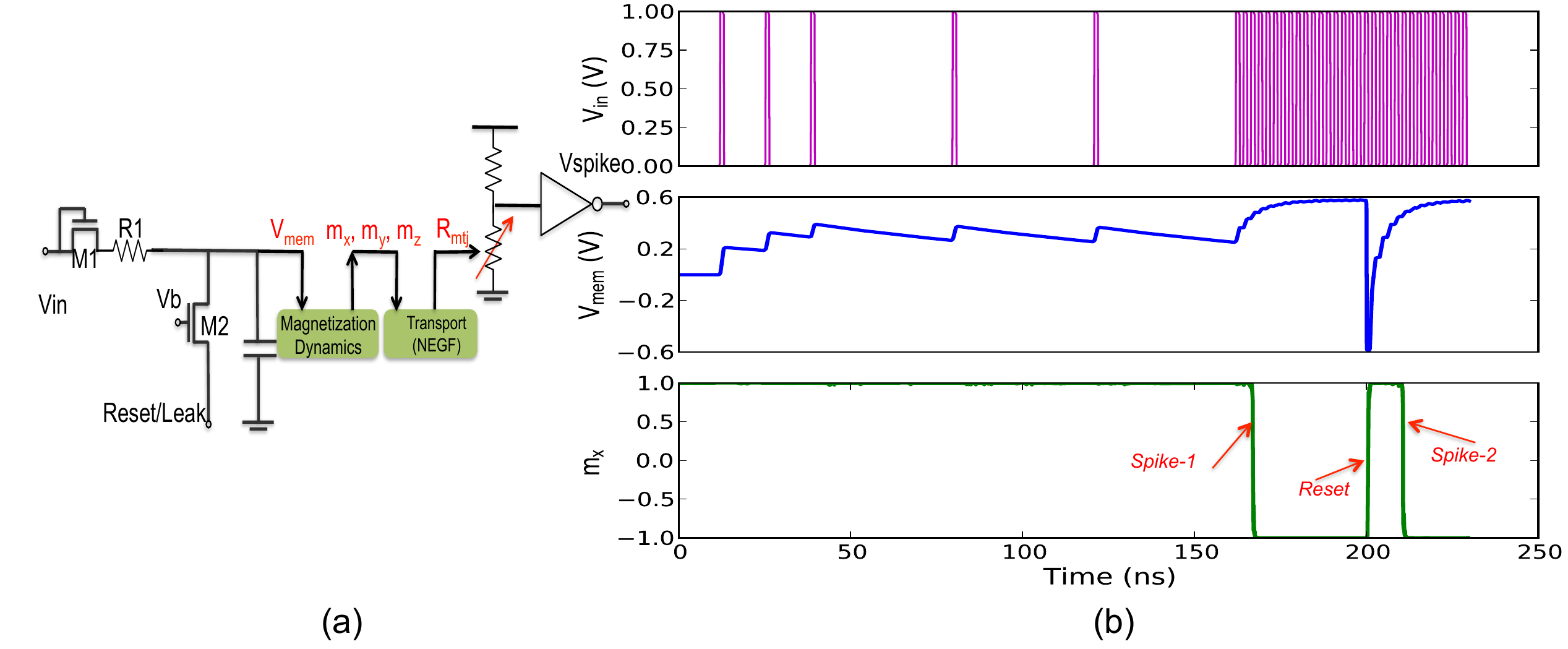}% Here is how to import EPS art
\caption{\label{fig:epsart} (a) Mixed mode simulation flow used in the present work. (b) Simulation results for the proposed ME device, shown in Fig. 2. Top panel shows the input spikes fed to the $V_{in}$ terminal of the device. Middle panel shows the voltage across the ME capacitor, exhibiting the typical leaky-integrate dynamics. Bottom panel, illustrates the switching of the ferro-magnet from +x to -x direction generating a spike annotated as \textit{Spike-1}. No more spikes are generated until the device is reset to its initial position by applying a negative voltage. After reset, device emits a second spike annotated as \textit{Spike-2}.}
\end{figure}

The evolution of the magnetization vector under the influence of an applied electric field was modeled using the well-know phenomenological equation known as the \textit{Landau-Lifshiz-Gilbert} (LLG) equation \cite{LLG}. LLG equation can be written as  
\begin{equation}
\label{LLG_eqn}
\frac{\partial \widehat{m}}{\partial \tau} = - \widehat{m} \times \vec{H}_{EFF} - \alpha \widehat{m} \times \widehat{m} \times \vec{H}_{EFF}  
\end{equation}
where $\tau$ is $\frac{\vert \gamma \vert }{1 + \alpha^2}t$. $\alpha$ is the Gilbert damping constant, $\gamma$ is the gyromagnetic ratio, $\widehat{m}$ is the unit magnetization vector, $t$ is time and $H_{EFF}$ is the effective magnetic field. $H_{EFF}$ includes the various fields acting on the magnet and can be written as
\begin{equation}
H_{EFF} = \vec{H}_{demag} + \vec{H}_{interface} + \vec{H}_{thermal} + \vec{H}_{ME}
\end{equation} 
where $\vec{H}_{demag}$ is the demagnetization field due to shape anisotropy. $\vec{H}_{interface}$ is interfacial perpendicular anisotropy, $\vec{H}_{thermal}$ is the stochastic field due to thermal noise and $\vec{H}_{ME}$ is the field due to ME effect.

In SI units $\vec{H}_{demag}$ can be expressed as
\begin{equation}
\label{H_demag}
\vec{H}_{demag} = - M_S(\, N_{xx}m_{x}\widehat{x},\,  N_{yy}m_{y}\widehat{y},\,  N_{zz}m_{z}\widehat{z} \,)
\end{equation}
where $m_{x}$, $m_{y}$ and $m_{z}$ are the magnetization moments in x, y and z directions respectively.  $N_{xx}$, $N_{yy}$ and $N_{zz}$ are the demagnetization factors for a rectangular magnet which can be estimated from analytical equations presented in \cite{LLG_f}. $M_s$ is the saturation magnetization. The interfacial perpendicular anisotropy can be represented as

\begin{table}[!t]
\renewcommand{\arraystretch}{1.7}
\centering
\caption{Summary of parameters used in our simulations}
\label{table1}

\begin{tabular}{c c}
\hline \hline
\bfseries Parameters & \bfseries Value\\
\hline
Magnet Length ($L_{mag}$) & $ 45nm \times 2.5 $\\
Magnet Width ($W_{mag}$) & $ 45nm$\\
Magnet Thickness ($t_{FL}$) & $ 2.5nm$\\
ME Oxide Length ($L_{ME}$) & $ 60nm$\\
ME Oxide Thickness ($t_{ME}$) & $ 5nm$\\
Saturation Magnetization ($M_{S}$) & 1257.3 $KA/m$ \cite{ikeda}\\
Gilbert Damping Factor ($\alpha$) & 0.03 \\
Interface Anisotropy ($K_{i}$) & $1 mJ/m^{2}$ \cite{ikeda}\\
ME Co-efficient ($\alpha_{ME}$) & $0.5/c* ms^{-1}$ \\
Relative Di-electric constant ($\epsilon_{ME}$) & 500 \cite{Intel_ME}\\
Temperature ($T$) & 300K \\
CMOS Technology & 45nm PTM \cite{PTM} \\

\hline \hline
$*c = Speed \ of \ light.$ 
\end{tabular}
\end{table}

\begin{equation}
\label{H_interface}
\vec{H}_{interface} =  (\, 0\widehat{x},\,  0\widehat{y},\, \frac{2K_{i}}{\mu_{o}M_St_{FL}} m_{z}\widehat{z} \,)  
\end{equation}
where $K_{i}$ is the effective energy density for interface perpendicular anisotropy and $t_{FL}$ is thickness of the free layer. The ME effect can be abstracted through the parameter $\alpha_{ME}$ \cite{Intel_ME} and can be written as
\begin{equation}
\vec{{H}}_{ME} = ( \alpha_{ME} (\frac{V_{ME}}{t_{ME}})\widehat{{x}},\,  0\widehat{{y}},\, 0\widehat{{z}} )
\end{equation}
where, $\alpha_{ME}$ is the co-efficient for ME effect, $V_{ME}$ is the voltage applied across the ME oxide and $t_{ME}$ is thickness of the ME oxide. $\vec{{H}}_{ME}$ is multiplied with a suitable constant for unit conversion.

To account for the effect of thermal noise, we included a stochastic field given by \cite{brown}
\begin{equation}
\label{H_thermal}
\vec{H}_{thermal} = \vec{\zeta}\sqrt{\frac{2\alpha k_BT}{\vert \gamma \vert M_SVol\ dt}}
\end{equation}
where $\vec{\zeta}$ is a vector with components that are zero mean Gaussian random variables with standard deviation of 1. $Vol$ is the volume of the nano-magnet, $T$ is ambient temperature, $dt$ is simulation time step and $k_B$ is Boltzmann's constant. 

The above set of stochastic differential equations (1)-(6) were integrated numerically by using the Heun's method. Device dimensions and other material parameters used in our simulations are mentioned in Table I. Fig. 3(a) illustrates a typical evolution of the components of magnetization vector under an applied electric field.

Additionally, the resistance of the MTJ stack was modeled using the non-equilibrium Green's function (NEGF) formalism. The results obtained from the NEGF equations were then abstracted into a behavioral model and used for SPICE simulations. Thus, a coupled device-to-circuits simulation framework was developed for analyzing the proposed ME neuron.

Fig. 3(b), shows the switching probability of the ferro-magnet as a function of the voltage on ME capacitor. The stochastic behavior of the switching mechanism can be attributed to the fluctuations in initail position of the magnetization direction due to thermal noise. The noisy characteristic of the proposed ME neuron mimics the stochasticity of biological neurons. In Fig. 4(b), we show the results from a mixed mode SPICE-MATLAB simulation of the device as shown in Fig. 4(a). As expected, the voltage on the ME capacitor, ($V_{mem}$ in Fig. 4(b)), shows the typical leaky-integrate behavior. If the accumulated voltage is sufficient enough, the device switches from +x to -x direction, governed by the magnetization dynamics equations. The output of the inverter goes high to produce a spike. The neuron (ferro-magnet) remains non-responsive to further input spikes, unless it is reset by applying a negative pulse on the \textit{Reset/Leak} terminal, as shown in Fig. 4(b).

As compared to a CMOS-only implementation, the non-volatile ferro-magnet could help reduce the leakage power of the neuronal circuit. Moreover, for a CMOS LIF neuron, the output spike has to be latched either to mimic the refractory period of the neuron or to wait for the peripheral hardware circuit to read the output spike and do necessary computations. In the proposed device, the latching operation is inherent in the ferro-magnet due to its non-volatility. Also, as compared to the recent experimental demonstration of an integrate-fire neuron in phase change device \cite{Ibm_pc}, the present proposal can potentially be  more energy-efficient due to lower operating voltages, and has the inherent benefit of almost unlimited endurance. In addition, the proposed ME device implements a leaky-integrate-fire neuron as opposed to the integrate-fire neuron of \cite{Ibm_pc}.

\section*{\large\bf{Spiking Neuromorphic Architecture}}
\subsection*{\normalsize\bf{SNN Topology for Pattern Recognition}}
\begin{figure}[!b]
\centering
\includegraphics[width=5.5in]{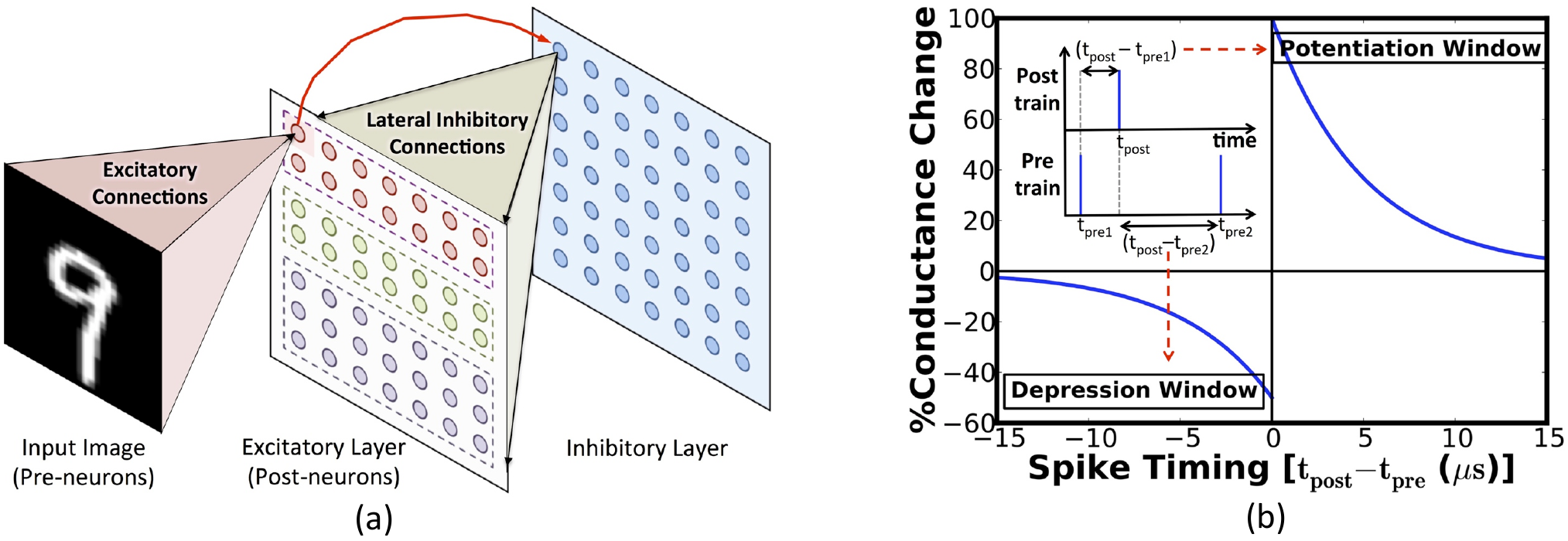}% Here is how to import EPS art
\caption{\textbf{(a)} SNN topology for pattern recognition. The input neurons are fully connected to the excitatory post-neurons, each of which is connected to the corresponding inhibitory neuron in a one-on-one manner. There are lateral inhibitory connections from each inhibitory neuron to all the excitatory post-neurons except the one from which it received a forward connection. \textbf{(b)} STDP learning algorithm, wherein the change in synaptic conductance is exponentially related to the difference in the spike times of the pre- and post-neuronal pair.}
\label{fig:SNN_Arch_STDP}
\end{figure}
We evaluate the applicability of the proposed neuron on a two-layered SNN used for pattern recognition as shown in Fig. \ref{fig:SNN_Arch_STDP}(a). Each pixel in the input image pattern constitutes an input neuron whose spike rate is proportional to the corresponding pixel intensity. The input pre-neurons are fully connected to every ME post-neuron in the excitatory layer. The excitatory post-neurons are further connected to the inhibitory neurons in a one-on-one manner, each of which inhibits all the excitatory neurons except the forward-connected one. Lateral inhibition prevents multiple post-neurons from spiking for similar input patterns. The excitatory post-neurons are further divided into various groups, where the neurons belonging to a group are trained to recognize varying representations of a particular class of input patterns fixed \textit{a priori}.

\subsection*{\normalsize\bf{Synaptic Learning Mechanism}}
The synapses connecting the input neurons to the post-neurons (excitatory connections in Fig. \ref{fig:SNN_Arch_STDP}(a)) are subjected to synaptic learning, which causes the connected post-neuron to spike exclusively for a class of input patterns. Spike timing dependent plasticity (STDP), wherein the synaptic conductance is updated based on the extent of temporal correlation between pre- and post-neuronal (post-synaptic) spike trains is widely used to achieve plasticity in SNNs. The strength of a synapse is increased/potentiated (decreased/depressed) if a pre-spike occurs prior to (later than) the post-spike as shown in Fig. \ref{fig:SNN_Arch_STDP}(b). The naive STDP algorithm considers the correlation only between pairs of pre- and post-synaptic spikes, while ignoring the information embedded in the post-neuronal spiking frequency. Hence, we explore an enhanced STDP algorithm, wherein the STDP-driven synaptic updates are regulated by a low-pass filtered version of the membrane potential \cite{STDP} that is a proxy for the post-neuronal spiking rate. According to the enhanced STDP algorithm, an STDP-driven synaptic update is carried out only if the filtered membrane potential of the corresponding post-neuron exceeds a pre-specified threshold. This ensures that synaptic learning is performed only on those synapses, where the connected post-neuron spikes at a higher rate indicating a strong correlation with the input pattern.

Additionally, we augmented the enhanced STDP algorithm with a reinforcement mechanism to further improve the efficiency of synaptic learning. According to this scheme, each post-neuron in the excitatory layer is designated \textit{a priori} to learn a specific class of input pattern. During the learning phase, the corresponding synapses are potentiated (depressed) if the post-neuron spikes for an input pattern whose class matches with (differs from) its designated class. The reinforced learning scheme enables the synapses to encode a better representation of the input patterns.

\subsection*{\normalsize\bf{Hardware Implementation}}
\begin{figure}[!t]
\centering
\includegraphics[width=4.0in]{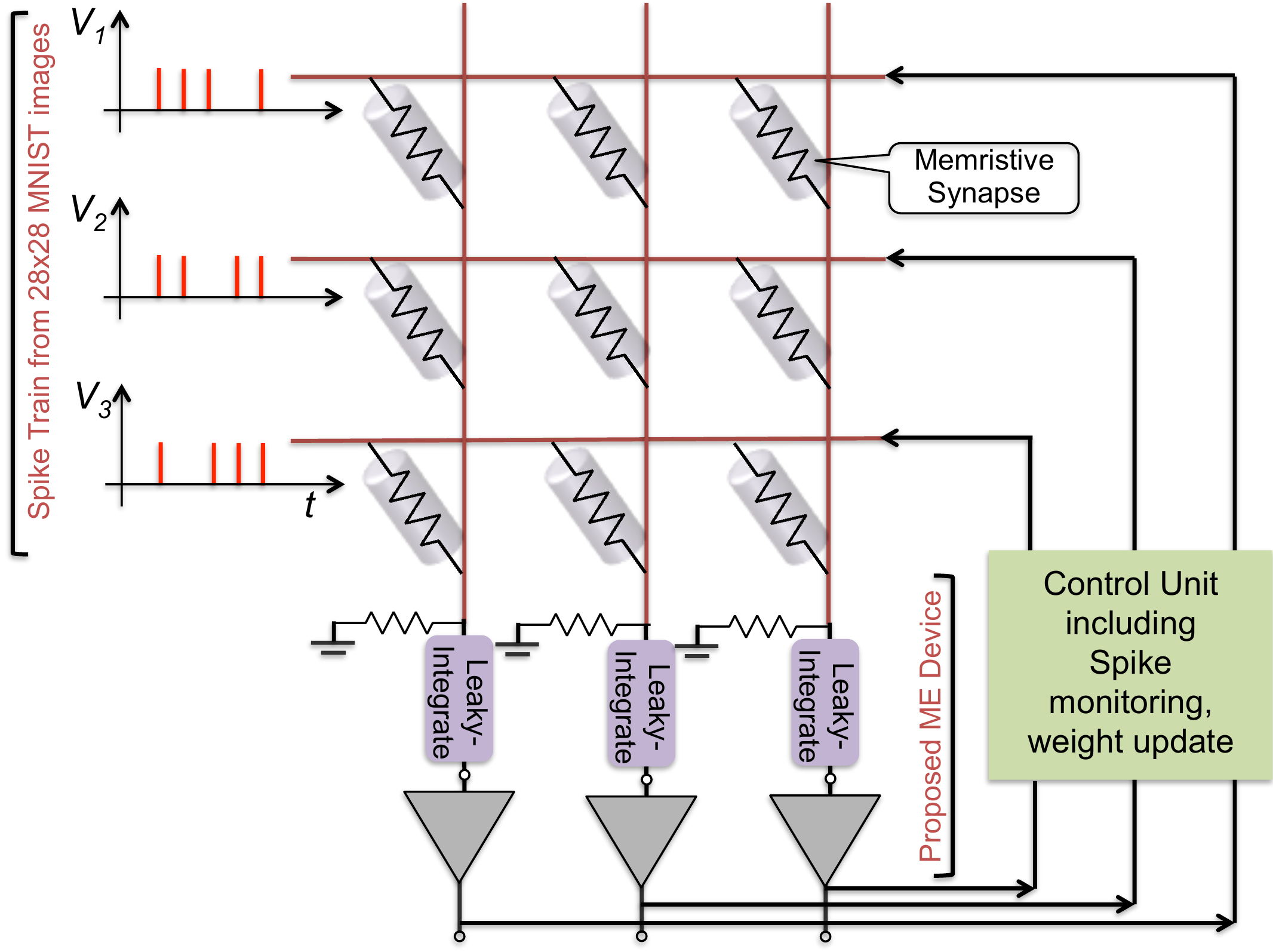}% Here is how to import EPS art
\caption{A typical crossbar implementation of the SNN topology using the proposed ME neuron. Memristive devices constitute the synapses, while the proposed device mimics the LIF post-neurons. The on-chip learning circuit programs the synaptic conductance based on spike timing. Inputs to the system are spike trains corresponding to the 28$\times$28 image pixels from the MNIST dataset.}
\label{fig:Crossbar_archi}
\end{figure}
We present a crossbar arrangement of the synapses and ME neurons (Fig. \ref{fig:Crossbar_archi}) for an energy-efficient realization of the SNN. Multilevel memristive technologies \cite{Nano_Memristor, Bipin} and spintronic devices \cite{Spin_Synapse} have been proposed to efficiently mimic the synaptic dynamics. Each pre-neuronal voltage spike is modulated by the interconnecting synaptic conductance to generate a resultant current into the ME neuron. The neuron integrates the current leading to an increase in its membrane potential, which leaks until the arrival of subsequent voltage spikes at the input. The ME neuron switches conditionally based on the membrane potential, to produce an output spike. The on-chip learning circuit samples the post-neuronal spike to program the corresponding synaptic conductances based on spike timing \cite{Bipin}. The energy-efficiency of the crossbar architecture stems from the localized arrangement of the neurons and synapses compared to von-Neumann machines with decoupled memory and processing units.

\subsection*{\normalsize\bf{Simulation Methodology}}
We developed a comprehensive device to system-level simulation methodology to evaluate the efficacy of an SNN composed of the proposed ME neurons for a pattern recognition application. The leaky-integrate-fire (LIF) dynamics of the ME neuron were validated using the mixed-mode device-circuit simulation framework as shown in Fig. 4(a). The crossbar architecture of a network of such ME neurons was simulated using an open-source SNN simulator known as BRIAN \cite{BRAIN} for recognizing digits from the MNIST dataset \cite{MNIST}. The leaky-integrate characteristics of the ME neurons were modeled using differential equations with suitable time constants while the switching dynamics were determined from stochastic LLG simulations. The synapses were modeled as multilevel weights. The enhanced STDP algorithm was implemented by recording the time instants of pre- and post-spikes, and regulating the weight updates with the averaged membrane potential.

Upon the completion of the training phase, digit recognition is performed by analyzing the spiking activity of different groups of neurons in the SNN, each of which learned to spike for a class of input patterns assigned \textit{a priori}. Each input image is predicted to represent the class (digit) associated with the neuronal group with the highest average spike count over the duration of the simulation. The classification is accurate if the actual class of the input image conforms to that predicted by the network of spiking neurons. The classification accuracy is then determined from the number of images correctly recognized by the SNN and the total number of input images. The classification performance is reported using ten thousand images from the MNIST testing image set. The read, reset, and ME capacitor charging energy consumed by the proposed neuron are estimated from SPICE simulations.

\section*{\large\bf{Results and Discussion}}
\begin{figure}[!t]
\centering
\includegraphics[width=5.5in]{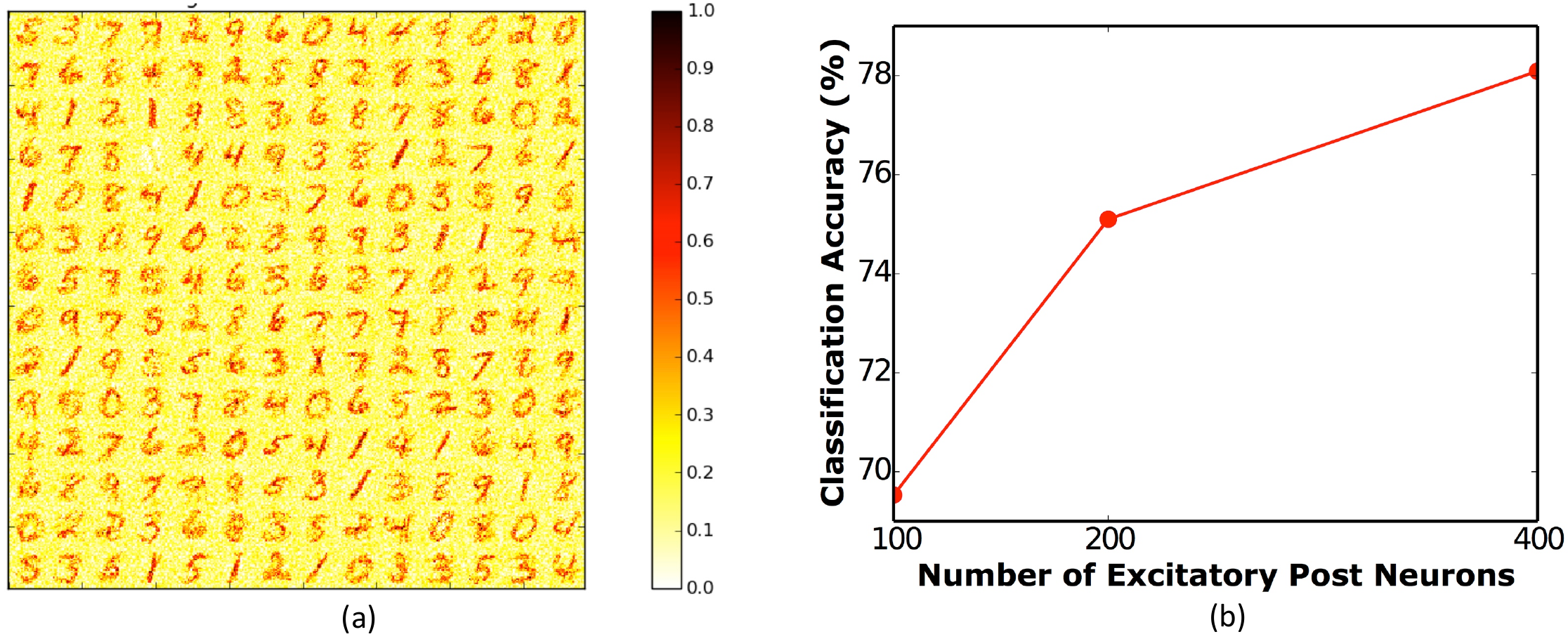}% Here is how to import EPS art
\caption{\textbf{(a)} Synaptic weights connecting the 28$\times$28 input pre-neurons to each of the 200 excitatory post-neurons towards the end of the training phase. \textbf{(b)} Classification accuracy verses the number of excitatory post-neurons.}
\label{fig:SNN_Results}
\end{figure}
Fig. \ref{fig:SNN_Results}(a) shows the synaptic weights connecting the 28$\times$28 input neurons to each of the 200 post-neurons towards the end of the training process. It can be seen that the synapses learned to encode the different input patterns. The LIF dynamics of the proposed ME neuron and the reinforced STDP learning algorithm helped achieve a classification accuracy of 70\% for a network of 100 neurons. It is evident from Fig. \ref{fig:SNN_Results}(b) that the classification performance can be improved by increasing the number of excitatory post-neurons. The proposed ME neuron consumed 17.5$fJ$ and 1.04$fJ$ for read and reset operations respectively. The average ME capacitor charging energy is estimated to be 246$fJ$ per neuron per training iteration, which is energy-efficient compared to CMOS neurons that were reported to consume $pJ$ of energy \cite{CMOS_Neuron}.   

\section*{\large\bf{Conclusion}}
Amid the quest for new device structures to mimic neuronal dynamics, we have proposed a spin based neuron using ME effect. As opposed to previous spiking neuron implementations in CMOS and other technologies, the proposed device combines low energy consumption and area efficiency along with the characteristic LIF dynamics of a biological neuron. Further, we believe, the similarity of the stochastic behavior of the proposed neuron with biological neurons, would open up new possibilities for efficient hardware implementations for a wider range of computational tasks.

\section*{\large\bf{Acknowledgments}}
The work was supported in part by, Center for Spintronic Materials, Interfaces, and Novel Architectures (C-SPIN), a MARCO and DARPA sponsored StarNet center, by the Semiconductor Research Corporation, the National Science Foundation, Intel Corporation and by the DoD Vannevar Bush Fellowship.

%\section*{\large\bf{Author Contributions}}
%A. Jaiswal and G. Srinivasan wrote the paper and performed the simulations. S. Roy assisted in performing the simulations. All authors helped with the writing of the paper, developing the concepts, and discussing the results.

%\section*{\large\bf{Competing Financial Interests}}
%The authors declare no competing financial interests.





%\bibliographystyle{IEEEtran}

\end{document}